\documentclass[12pt]{article}
\usepackage[french,english]{babel}
\usepackage{techreport}
\usepackage{graphicx}
\usepackage{amsmath}
\usepackage{amssymb}
\usepackage{amsfonts}
\usepackage{setspace}
\usepackage{natbib}
\usepackage{algorithm}
\usepackage{subfig}
\usepackage{mycommands}

\begin{document}
\title{Online algorithms for Nonnegative Matrix Factorization  with the Itakura-Saito divergence}

\author{\name Augustin Lef\`evre \email augustin.lefevre@inria.fr \\ 
\indent\addr INRIA/ENS Sierra project \\
\indent23, avenue d'Italie, 75013 Paris 
\AND
\name Francis Bach \email francis.bach@ens.fr \\
\indent\addr INRIA/ENS Sierra project \\
\indent 23, avenue d'Italie, 75013 Paris 
\AND
\name C\'edric F\'evotte \email fevotte@telecom-paristech.fr \\
\indent \addr CNRS LTCI; T\'el\'ecom ParisTech \\
\indent37-39, rue Dareau, 75014 Paris
}
\editor{}
\maketitle

\abstract{
Nonnegative matrix factorization (NMF) is now a common tool for audio source separation.\,When learning NMF on large audio databases, one major drawback is that the complexity in time is $O(F K N)$ when updating the dictionary  
(where $(F, N)$ is the dimension of the input power spectrograms, and $K$ the number of basis spectra), thus forbidding its application on signals longer than an hour.
We provide an online algorithm with a complexity of $O(F K)$ in time and memory for updates in the dictionary. 
We show on audio simulations that the online approach is faster for short audio signals and allows to analyze audio signals of several hours.
} 

\section{Introduction}

In audio source separation, nonnegative matrix factorization (NMF) is a popular technique for building a high-level representation of complex audio signals with a small number of basis spectra, forming together a dictionary \citep{plca_07,IS-NMF,tuomas-icassp}. 
Using the Itakura-Saito divergence as a measure of fit of the dictionary to the training set allows to capture fine structure in the power spectrum of audio signals as shown in \citep{IS-NMF}. 

However, estimating the dictionary can be quite slow for long audio signals, and indeed intractable for training sets of more than a few hours.
We propose an algorithm to estimate Itakura-Saito NMF (IS-NMF) on audio signals of possibly infinite duration with tractable memory and time complexity.
This article is organized as follows :
in Section \ref{sec:algo}, we summarize Itakura-Saito NMF, propose an algorithm for online NMF, and discuss implementation details. 
In Section \ref{sec:xp}, we experiment our algorithms on real audio signals of short, medium and long durations. 
We show that our approach outperforms regular batch NMF in terms of computer time. 

\section{Online Dictionary Learning for the Itakura-Saito divergence}
\label{sec:algo}

Various methods were recently proposed for online dictionary learning \citep{mairal10a,online_lda,bucak09}. 
However, to the best of our knowledge, no algorithm exists for online dictionary learning with the Itakura-Saito divergence.  
In this section we summarize IS-NMF, then introduce our algorithm for online NMF and explain briefly the mathematical framework.

\subsection{Itakura-Saito NMF}
\label{sec:bnmf}

Define the Itakura-Saito divergence as $d_{IS}(y,x)=\sum_i (\frac{y_i}{x_i} -\log \frac{y_i}{x_i} -1 )$. 
Given a data set $V = (v_1, \dots, v_N) \in \R_+^{F \times N}$, Itakura-Saito NMF consists in finding $ W \in \R_+^{F \times K} $, $ H =(h_1, \dots, h_N) \in \R_+^{K \times N} $ that minimize the following objective function : 

\begin{equation}
\mathcal{L}_{H}(W)= \frac{1}{N} \sum_{n=1}^N  d_{IS} ( v_n, W h_n) \, , 
\label{is_nmf}
\end{equation}
The standard approach to solving IS-NMF is to optimize alternately in $W$ and $H$ and use majorization-minimization~\citep{beta_nmf}. 
At each step, the objective function is replaced by an auxiliary function of the form $\mathcal{L}_H(W,\ul{W})$ such that $ \mathcal{L}_H(W) \leq \mathcal{L}_H(W,\ul{W})$ with equality if $W=\ul{W}$ :
\begin{equation}
 \mathcal{L}_H(W,\ul{W}) = \sum_{f k}  A_{f k} \frac{1}{W_{f k}} + B_{f k} W_{f k} + c\, .
\label{aux_fun0}
\end{equation}
where $A, B \in \R_+^{F \times K}$ and $c \in \R$ are given by: 
\begin{equation}
\begin{array}{rl} 
A_{f k}	&= \sum_{n=1}^N H_{k n} V_{f n} (\ul{W} H)^{-2}_{f n} \ul{W}_{f k}^2 \, ,\\
\\
B_{f k}	&= \sum_{n=1}^N H_{k n} (\ul{W} H)^{-1}_{f n} \, , \\
\\
c	&= \sum_{f=1}^F \sum_{n=1}^N \log \dfrac{\displaystyle V_{f n}}{(\ul{W} H)_{f n}} - F \, .   
\end{array} 
\label{aux_fun}
\end{equation}
Thus, updating  $W$ by $W_{f k} =\sqrt{A_{f k}/B_{f k}}$ yields a descent algorithm. 
Similar updates can be found for $h_n$ so that the whole process defines a descent algorithm in $(W,H)$ (for more details see, e.g.,~\citep{beta_nmf}). In a nutshell, batch IS-NMF works in cycles: at each cycle, all sample points are visited, the whole matrix $H$ is updated, the auxiliary function in \equ{aux_fun0} is re-computed, and $W$ is then updated. We now turn to the description of online NMF.
\subsection{An online algorithm for online NMF}
\label{sec:onmf}

When $N$ is large, multiplicative updates algorithms for IS-NMF become expensive because at the dictionary update step, they involve large matrix multiplications with time complexity in $O(F K N)$ (computation of matrices $A$~and~$B$). 
We present here an online version of the classical multiplicative updates algorithm, in the sense that only a subset of the training data is used at each step of the algorithm. 

Suppose that at each iteration of the algorithm we are provided a new data point $v_t$, and we are able to find $h_t$ that minimizes  $d_{IS}(v_t, W^{(t)} h_t)$. 
Let us rewrite the updates in \equ{aux_fun}. 
Initialize $A^{(0)},B^{(0)},W^{(0)}$ and at each step compute~:
\begin{equation}
\begin{array}{rl}
A^{(t)} = & A^{(t-1)} + (\frac{v_t}{(W^{(t-1)} h_t)^{2}} h_t^\top) \cdot (W^{(t-1)})^2 \, ,\\
\\
B^{(t)} = & B^{(t-1)} + \frac{1}{W^{(t-1)} h_t} h_t^\top  \, ,\\
\\
W^{(t)} = & \sqrt{\frac{A^{(t)}}{B^{(t)}}} \, .			\\
\end{array}
\end{equation}
Now we may update $W$ each time a new data point $v_t$ is visited, instead of visiting the whole data set. 
This differs from batch NMF in the following sense~: suppose we replace the objective function in \equ{is_nmf} by 

\begin{equation}
L_T(W)= \frac{1}{T} \sum_{t=1}^T d_{IS}(v_t, W h_t) \, , \\
\label{online_loss}
\end{equation}
where $(v_1, v_2, \dots, v_t, \dots)$ is an infinite sequence of data points, and the sequence $( h_1, \dots, h_t,\dots )$ is such that $h_t$ minimizes  $d_{IS}(v_t, W^{(t)} h)$.
Then we may show that the modified sequence of updates corresponds to minimizing the following auxiliary function : 
\begin{equation}
\hat{L}_T(W)= \sum_k \sum_f \left( A^{(T)}_{f k} \frac{1}{W_{f k}} + B^{(T)}_{f k} W_{f k}\right) +c \, .
\end{equation} 
If $T$ is fixed, this problem is exactly equivalent to IS-NMF on a finite training set. Whereas in the batch algorithm described in Section \ref{sec:bnmf}, all $H$ is updated once and then all $W$, in online NMF, each new $h_t$ is estimated exactly and then $W$ is updated once. 
Another way to see it is that in standard NMF, the auxiliary function is updated at each pass through the whole dataset from the most recent updates in $H$, whereas in online NMF, the auxiliary function takes into account all updates starting from the first one. 

\paragraph{Extensions} Prior information on $H$ or $W$ is often useful for imposing structure in the factorization \citep{gis-nmf,nmf_virtanen,plca_07}. Our framework for online NMF easily accomodates penalties such as :
\begin{itemize}
\item Penalties depending on the dictionary $W$ only.
\item Penalties on $H$ that are decomposable and expressed in terms of a concave increasing function $\psi$ \citep{gis-nmf}: $\Psi(H)=\sum_{n=1}^N \psi(\sum_k H_{k n})$.
\end{itemize}

\subsection{Practical online NMF}
\label{sec:extensions}

\begin{algorithm}
\caption{Online Algorithm for IS-NMF}
 \begin{tabular}{l}
\textbf{Input} training set, $W^{(0)}$, $A^{(0)}$, $B^{(0)}$, $\rho$, $\beta$, $\eta$, $\varepsilon$. \\
$t \leftarrow 0$ \\
  \textbf{repeat}  \\
 \qquad $t \leftarrow t +1$ \\ 
 \qquad draw $v_t$ from the training set. \\ 
 \qquad $h_t \leftarrow \mathop{\arg\min}_{h} d_{IS}(\varepsilon + v_t, \varepsilon + W h)$ \\ 
 \qquad  $ a^{(t)} \leftarrow (\frac{\varepsilon + v_t}{(\varepsilon + W h_t)^{2}} h_t^\top) \cdot W^2$ \\
 \qquad  $ b^{(t)} \leftarrow \frac{1}{\varepsilon + W h_t} h_t^\top$ \\
 \qquad \textbf{if} $t \equiv 0 \;[\beta]$ \\
 \qquad \qquad $A^{(t)} \leftarrow A^{(t-\beta)} + \rho \sum_{s=t-\beta+1}^t a^{(s)}$ \\
 \qquad \qquad $B^{(t)} \leftarrow B^{(t-\beta)} + \rho \sum_{s=t-\beta+1}^t b^{(s)}$ \\
 \qquad \qquad  $ W^{(t)} \leftarrow \sqrt{\frac{A^{(t)}}{B^{(t)}}} $ \\
 \qquad \qquad \textbf{for} $k=1 \dots K$ \\
 \qquad \qquad \qquad $s \leftarrow \sum_f W_{f k} \, ,\quad W_{f k} \leftarrow W_{f k}/s $\\
 \qquad \qquad \qquad $A_{f k} \leftarrow A_{f k}/s \, ,\quad B_{f k} \leftarrow B_{f k} \times s $\\
 \qquad \qquad \textbf{end for} \\
\qquad \textbf{end if} \\
 \textbf{until} $\| W^{(t)} - W^{(t-1)} \|_F < \eta$  \\
 \end{tabular}
\label{alg:onmf}
\end{algorithm}
We provided a description of a pure version of online NMF, we now discuss several extensions that are commonly used in online algorithms and allow for considerable gains in speed. 

\paragraph{Finite data sets.}
When working on finite training sets, we cycle over the training set several times, and randomly permute the samples at each cycle. 

\paragraph{Sampling method for infinite data sets.}
When dealing with large (or infinite) training sets, samples may be drawn in batches and then permuted at random to avoid local correlations of the input.

\paragraph{Fresh or warm restarts.} 
Minimizing $d_{IS}(v_t,W h_t)$ is an inner loop in our algorithm. Finding an exact solution $h_t$ for each new sample may be costly (a rule of thumb is $100$ iterations from a random point). 
A shortcut is to  stop the inner loop before convergence. This amounts to compute only an upper-bound of $d_{IS}(v_t, W h_t)$.
Another shortcut is to warm restart the inner loop, at the cost of keeping all the most recent regression weights $H=(h_1, \dots, h_N)$ in memory. 
For small data sets, this allows to run online NMF very similarly to batch NMF : each time a sample is visited $h_t$ is updated only once, and then $W$ is updated. When using warm restarts, the time complexity of the algorithm is not changed, but the memory requirements become $O((F + N)K)$. 

\paragraph{Mini-batch.} 
Updating $W$ every time a sample is drawn costs $O(F K)$ : as shown in simulations, we may save some time by updating $W$ only every $\beta$ samples i.e., draw samples in batches and then update $W$. This is also meant to stabilize the updates.

\paragraph{Scaling past data.}
In order to speed up the online algorithm it is possible to scale past information so that newer information is given more importance :

\begin{equation}
\begin{array}{cc}
A^{(t+\beta)} &= A^{(t)} + \rho \sum_{s=t+1}^{t+\beta} a^{(s)} \, , \\
B^{(t+\beta)} &= B{(t)} + \rho \sum_{s=t+1}^{t+\beta} b^{(s)} \, , \\
\end{array}
\end{equation}
where we choose $\rho=r^{\beta/N}$. We choose this particular form so that when $N \rightarrow + \infty$, $\rho=1$. 
Moreover, $\rho$ is taken to the power $\beta$ so that we can compare performance for several batch sizes and the same parameter $r$. 
In principle this rescaling of past information amounts to discount each new sample at rate $\rho$, thus replacing the objective function in \equ{online_loss} by :
\begin{equation}
\frac{1}{\sum_{t=1}^T r^t} \sum_{t=1}^T r^{T+1-t} l(v_t, W) \, , \\
\end{equation}

\paragraph{Rescaling $W$.} 
In order to avoid the scaling ambiguity, each time $W$ is updated, we rescale $W^{(t)}$ so that its columns have unit norm. 
$A^{(t)}$, $B^{(t)}$ must be rescaled accordingly (as well as $H$ when using warm restarts). 
This does not change the result and avoids numerical instabilities when computing the product $W H$.

\paragraph{Dealing with small amplitude values.} 
The Itakura-Saito divergence $d_{IS}(y,x)$ is badly behaved when either $y=0$ or $x=0$. 
As a remedy we replace it in our algorithm by $d_{IS}(\varepsilon + y ,\varepsilon + x )$. The updates were modified consequently in Algorithm \ref{alg:onmf}.

\paragraph{Overview.}
Algorithm \ref{alg:onmf} summarizes our procedure. 
The two parameters of interest are the mini-batch size $\beta$ and the forgetting factor $r$. 
Note that when $\beta = N$, and $r=0$, the online algorithm is equivalent to the batch algorithm. 

\section{Experimental study}
\label{sec:xp}

In this section we validate the online algorithm and compare it with its batch counterpart. 
A natural criterion is to train both on the same data with the same initial parameters $W^{(0)}$ (and $H^{(0)}$ when applicable) and compare their respective fit to a held-out test set, as a function of the computer time available for learning. 
The input data are power spectrogram extracted from single-channel audio tracks, with analysis windows of $512$ samples and $256$ samples overlap. 
All silent frames were discarded.

We make the comparison for small, medium, and large audio tracks (resp. $10^3,\, 10^4, 10^5$ time windows).
$W$ is initialized with random samples from the train set. 
For each process, several seeds were tried, the best seed (in terms of objective function value) is shown for each process.
Finally, we use $\varepsilon=10^{-12}$ which is well below the hearing threshold.

\begin{figure}[htdp]
\centering
\begin{tabular}{cc}
\includegraphics[scale=0.4]{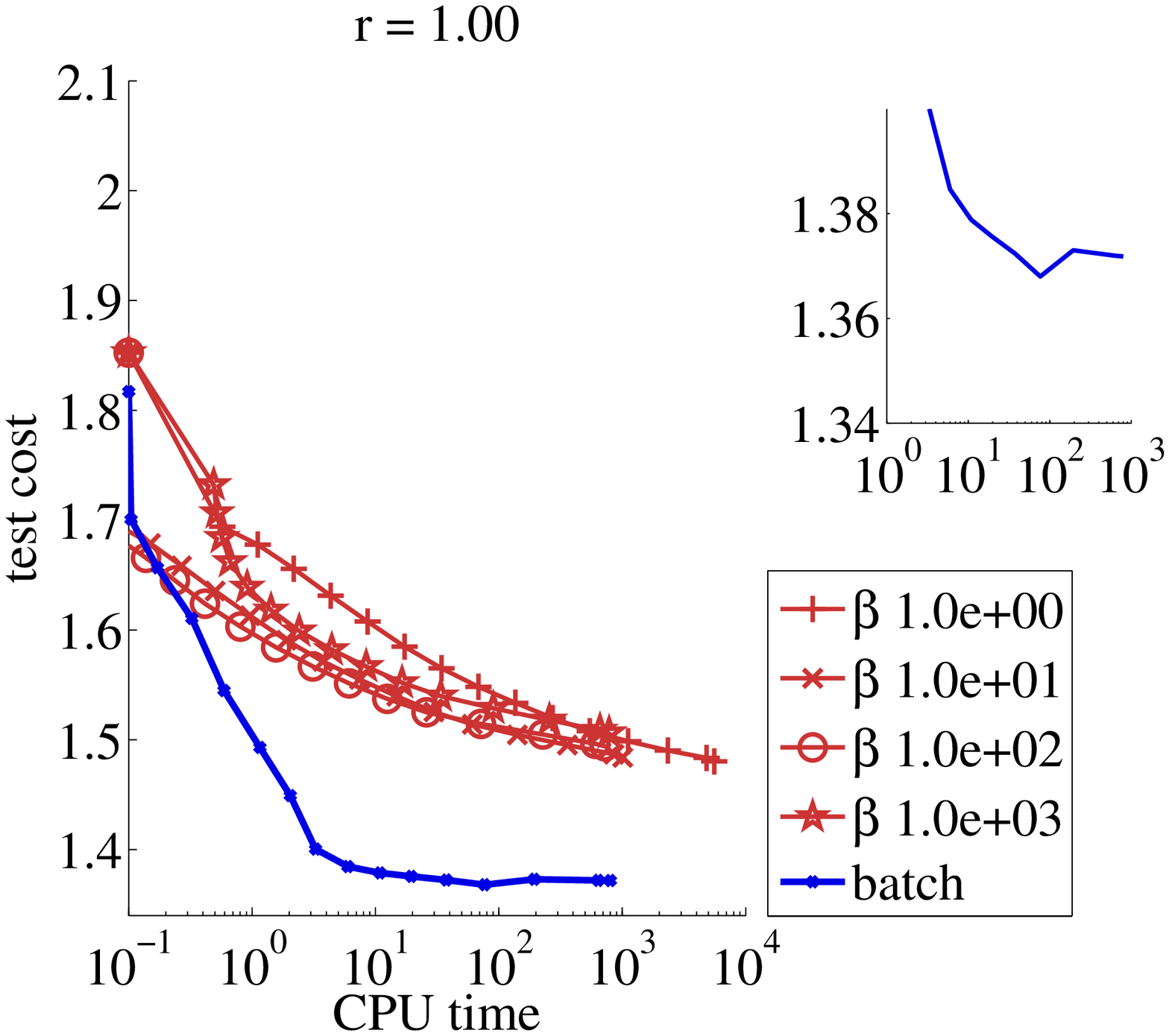} & 
\includegraphics[scale=0.4]{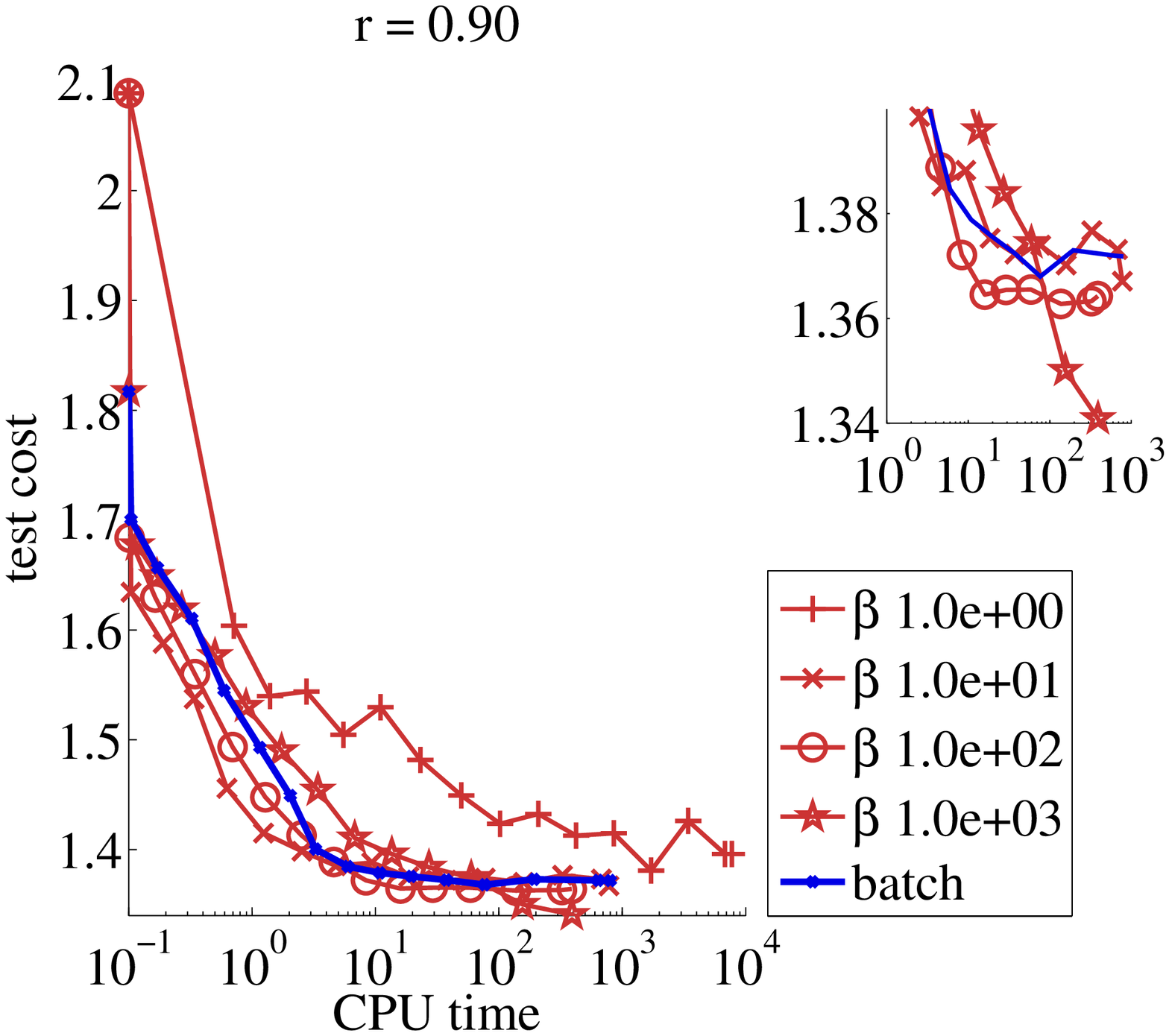} \\ 
\includegraphics[scale=0.4]{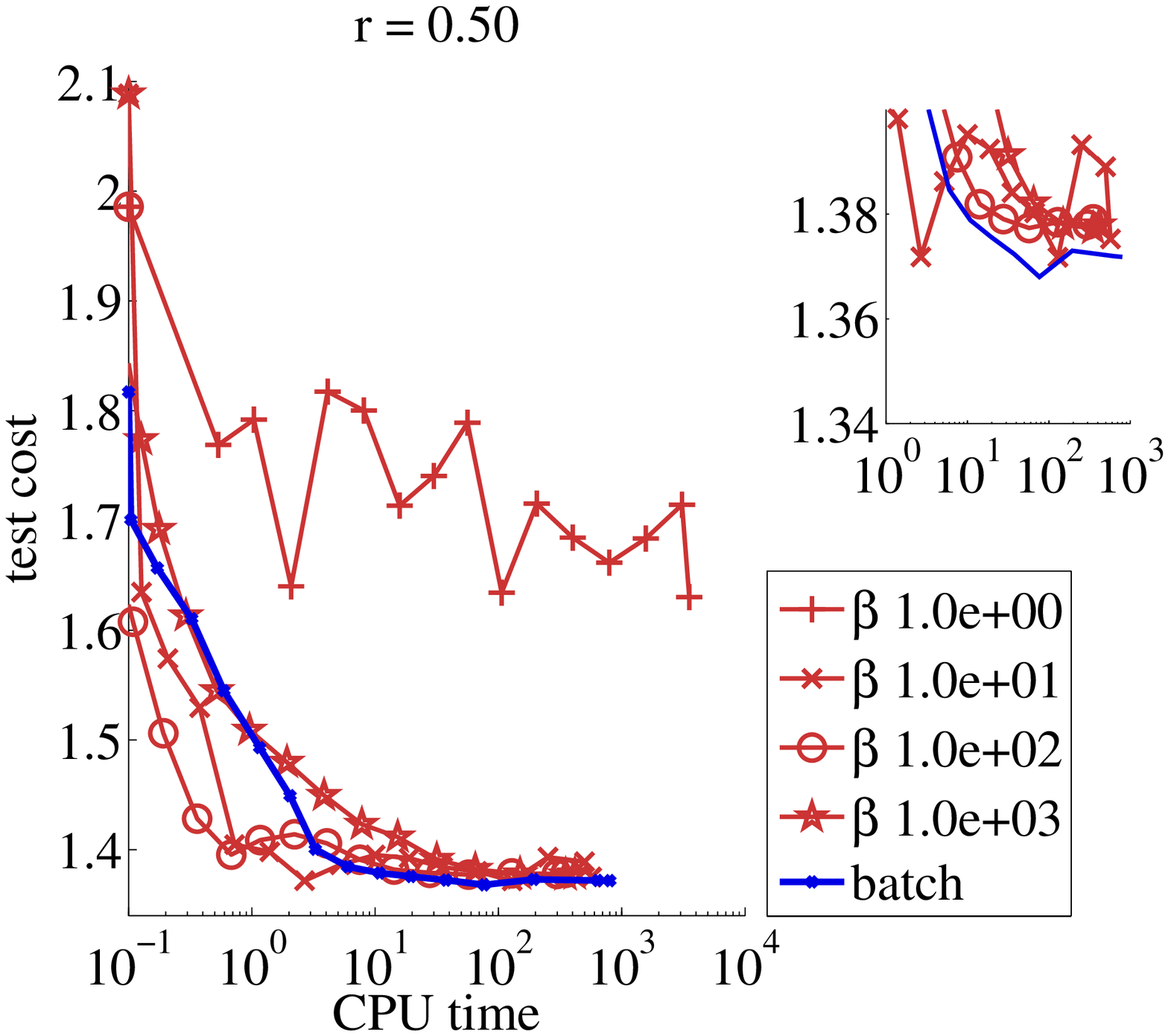}
\end{tabular}
\caption{Comparison of online and batch algorithm on a thirty-seconds long audio track.}
\label{tpms}
\end{figure}

\paragraph{Small data set (30 seconds).} We ran online NMF with warm restarts and one update of $h$ every sample.
From Figure \ref{tpms}, we can see that there is a restriction on the values of $(\beta,r)$ that we can use : if $r<1$ then $\beta$ should be chosen larger than~$1$. 
On the other hand, as long as $r >0.5$, the stability of the algorithm is not affected by the value of~$\beta$. 
In terms of speed, clearly setting $r<1$ is crucial for the online algorithm to compete with its batch counterpart.
Then there is a tradeoff to make in $\beta$ : 
it should picked larger than $1$ to avoid instabilities, and smaller than the size of the train set for faster learning (this was also shown in \citep{mairal10a} for the square loss). 

\begin{figure}[htdp]
\centering
\begin{tabular}{ccc}
\includegraphics[scale=0.4]{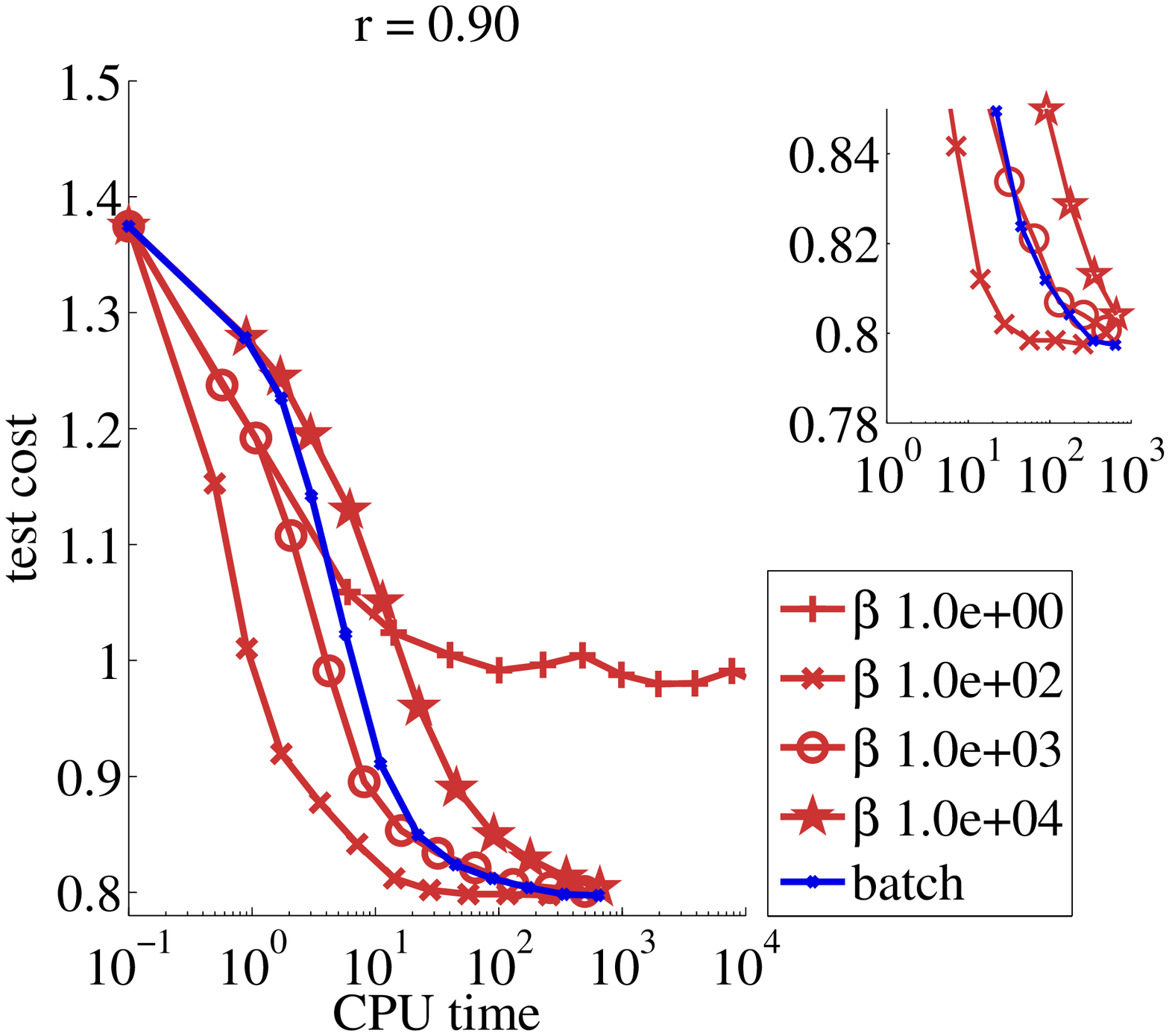} &
\includegraphics[scale=0.4]{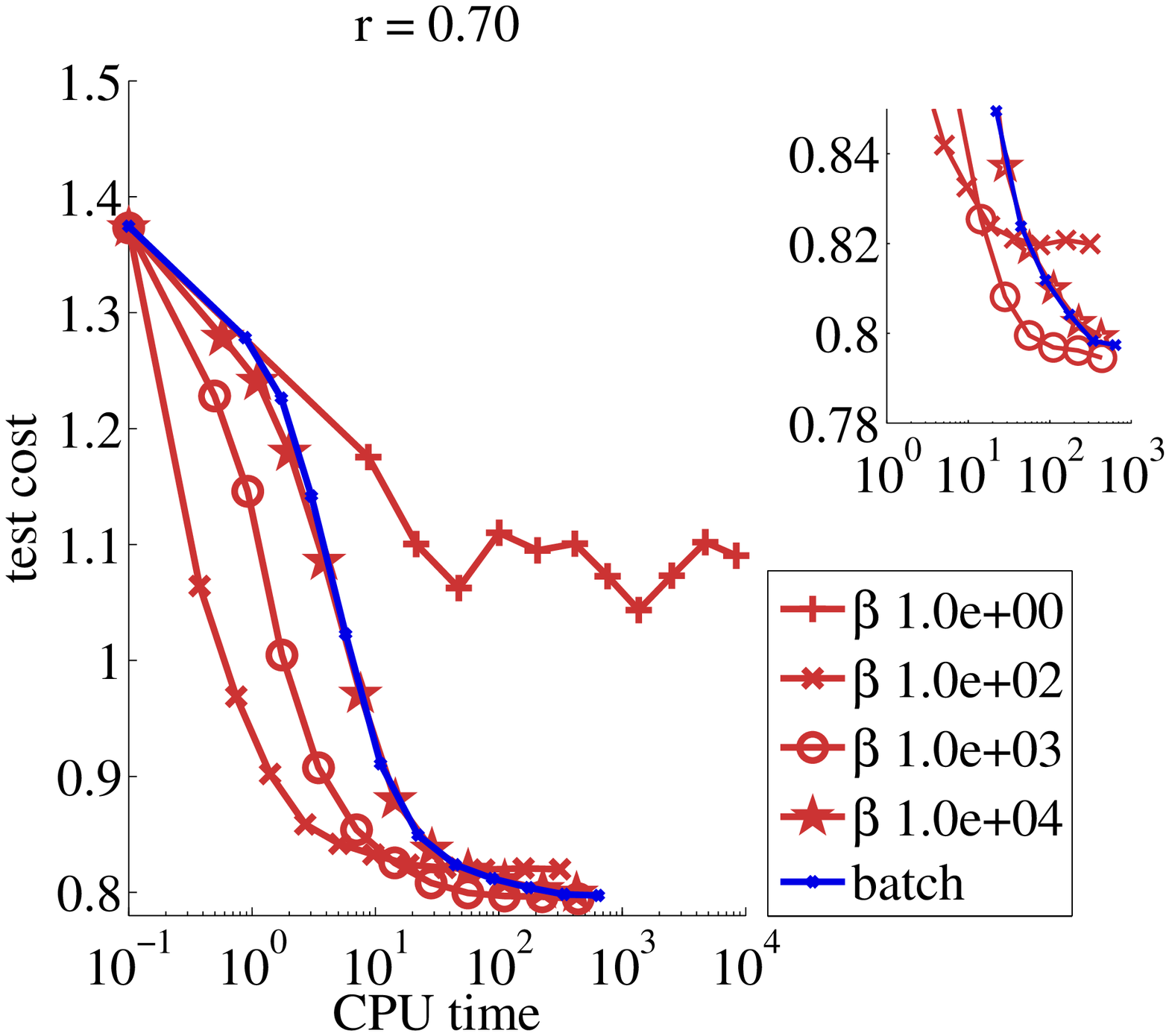} \\
\includegraphics[scale=0.4]{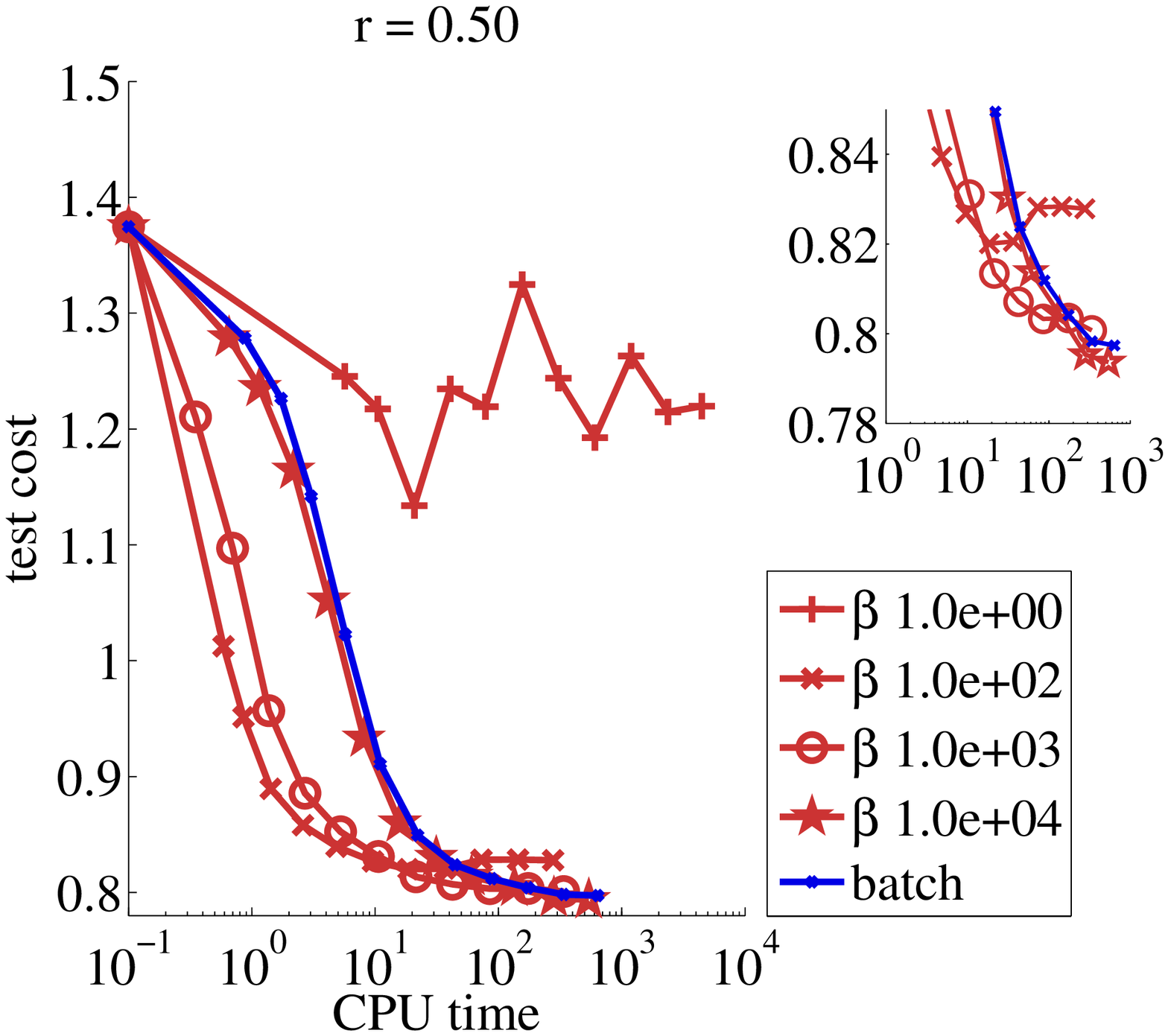}
\end{tabular}
\caption{Comparison of online and batch algorithm on a three-minutes long audio track.}
\label{hsv}
\end{figure}

\paragraph{Medium data set (4 minutes).} We ran online NMF with warm restarts and one update of $h$ every sample. 
The same remarks apply as before, moreover we can see on Figure \ref{hsv} that the online algorithm outperforms its batch counterpart by several orders of magnitude in terms of computer time for a wide range of parameter values.

\begin{figure}[htdp]
\centering
\includegraphics[scale=0.4]{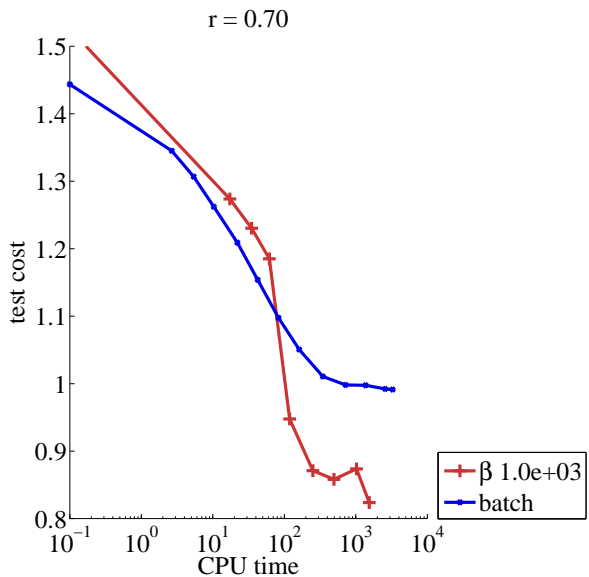}
\caption{Comparison of online and batch algorithm on an album of Django Reinhardt (1 hour 20 minutes).}
\label{dj}
\end{figure}

\paragraph{Large data set (1 hour 20 minutes).}For the large data set, we use fresh restarts and $100$ updates of $h$ for every sample. 
Since batch NMF does not fit into memory any more, we compare online NMF with batch NMF learnt on a subset of the training set.
In Figure \ref{dj}, we see that running online NMF on the whole training set yields a more accurate dictionary in a fraction of the time that batch NMF takes to run on a subset of the training set. We stress the fact that we used fresh restarts so that there is no need to store $H$ offline.

\paragraph{Summary.} \hspace{-.1cm}The online algorithm we proposed is stable provided minimal restrictions on the values of the parameters $(r, \beta)$ : if $r=1$, then any value of $\beta$ is stable. If $r<1$ then $\beta$ should be chosen large enough.
Clearly there is a tradeoff in choosing the mini-batch size $\beta$, which is explained by the way it works : 
when $\beta$ is small, frequent updates of $W$ are an additional cost as compared with batch NMF. 
On the other hand, when $\beta$ is small enough we take advantage of the redundancy in the training set.
From our experiments we find that choosing $r =0.7$ and $\beta=10^3$ yields satisfactory performance.

\section{Conclusion}

In this paper we make several contributions : 
we provide an algorithm for online IS-NMF with a complexity of $O(F K)$ in time and memory for updates in the dictionary.
We propose several extensions that allow to speedup online NMF and summarize them in a concise algorithm (code will be made available soon). 
We show that online NMF competes with its batch counterpart on small data sets, while on large data sets it outperforms it by several orders of magnitude. 
In a pure online setting, data samples are processed only once, with constant time and memory cost. 
Thus, online NMF algorithms may be run on data sets of potentially infinite size which opens up many possibilities for audio source separation. 

\selectlanguage{english}
\bibliographystyle{plainnat}
\bibliography{bib_online_nmf1}
\end{document}